# Finger Vein Recognition by Generating Code

Zhongxia Zhang, Mingwen Wang

## Abstract

Finger vein recognition has drawn increasing attention as one of the most popular and promising biometrics due to its high distinguishes ability, security and non-invasive procedure. The main idea of traditional schemes is to directly extract features from finger vein images or patterns and then compare features to find the best match. However, the features extracted from images contain much redundant data, while the features extracted from patterns are greatly influenced by image segmentation methods. To tack these problems, this paper proposes a new finger vein recognition by generating code. The proposed method does not require an image segmentation algorithm, is simple to calculate and has a small amount of data. Firstly, the finger vein images were divided into blocks to calculate the mean value. Then the centrosymmetric coding is performed by using the generated eigenmatrix. The obtained codewords are concatenated as the feature codewords of the image. The similarity between vein codes is measured by the ratio of minimum Hamming distance to codeword length. Extensive experiments on two public finger vein databases verify the effectiveness of the proposed method. The results indicate that our method outperforms the state-of-theart methods and has competitive potential in performing the matching task.

## 1 Introduction

Finger vein recognition is a novel physiological biometric for human recognition that uses vascular pattern underneath the skin on the finger palmar side to authenticate the personal identity[1-3]. There are several advantages to using a finger vein image for personal authentication compared to other biometric technologies:(1) safety: vein pattern is internal features, not easy to replicate [4]. (2) Living body identification: only the vein in a living finger can be captured and further used in identification. (3)Non-contact：not affected by skin conditions;

Like other biometric verification methods, a finger-vein based verification method mainly has four steps, data acquisition, data preprocessing, feature extraction, and data verification. In recent years, a variety of feature extraction methods for finger vein recognition have been proposed, which can be roughly divided into the following categories:

Vein pattern based methods segment the vein pattern from finger vein image, and the geometric shape or topological structure of vein patterns are used for matching, such as repeated line tracking (RLT) [5], maximum curvature [6], mean curvature [7], Gabor filter [8], and so on. The method of repeated line tracking was proposed by Miura [5]. Finger vein pattern is the characteristic of finger vein image itself. It is a linear vein existing according to the law, but the noise is irregular. Therefore, we can find out the vein line of finger through multiple tracking and filter out the interference in the image. But the disadvantages are high time complexity, not robust to large noise, can not effectively extract finger vein veins in areas with severe shadow, and do not consider the symmetry and continuity of veins. Moreover, Miura. et al. proposed the local maximum curvature method to extract vein features in [6]. Find the point of maximum curvature from four directions, and finally calculate the weighted sum of all points. Subsequently, both Literature [7] and [9] also used curvature method to extract finger vein features. Literature [8] used Gabor filter to extract vein patterns. Gabor filter's selectivity to center frequency and direction makes it powerful in image texture analysis. However, the Gabor filter will cause information loss for small and fuzzy veins in low-quality vein images. All of the above methods need to segment veins. Finger vein images are collected under near-infrared light, and the contrast of finger vein images is often low because of light attenuation caused by absorption of other tissues of fingers. Therefore, the performance of finger vein method which depends on segmentation results will degrade due to the degradation of image quality.

The method based on feature points matching: detect the feature points of the image, and use the feature points to match. This method can be divided into two categories: method based on minutiae and method based on SIFT feature. Minutiae includes bifurcation points and end points. Typical methods based on minutiae include minutiae matching based on improved Hausdorff distance matching [10]and minutiae matching based on singular value decomposition [11]. The minutiae based method needs to segment the finger vein like the vein pattern based method, and then extract the minutiae from the texture. In finger vein image, the number of minutiae is small, which is a problem in the application of minutiae based method to the recognition task of finger vein image. The SIFT [12,13] method can extract more feature points from finger vein images. However, the fuzzy vein patterns of finger vein images can easily lead to false detection of feature points, and the deformation of vein lines caused by finger bending or rotation is not considered.

The basic principle of statistical characteristics analysis based schemes such as principal component analysis (PCA) [14], linear discriminant analysis (LDA) [15] and sparse representation (SR) [16] is to transform the vein texture image into different subspaces, and generate feature vectors from various coefficients of the subspace to complete vein recognition. Wang et al. [14] combined the traditional 2D-PCA and 2D-FLD (Fisher linear discriminant) technology, Wu and Liu [15] used PCA and LDA to achieve finger vein classification, while Xin et al. [16] successfully applied SR to finger vein recognition tasks. These PCA, LDA and SR based methods can reduce the preprocessing steps and have small space occupation of feature vectors. However, they extract features from a global perspective and insufficiently describe local feature information. And these methods are not robust to local appearance variations that are caused by skin scattering, bone occlusion, uneven illumination, the pose and rotation of finger when imaging.

The method based on local features is widely used in finger vein recognition[17-19]. These methods include local binary mode (LBP) [20], local derivative mode (LDP) [21]. Many LBP variants have also been proposed. In [22], the author proposed a new LBP variant, called directional binary code. In [23], the author proposed to use personalized best bit mapping based on local binary pattern (LBP). Experiments showed that this feature not only has better performance, but also has higher robustness and reliability. Recently, Petpon and Srisuk [24] proposed a new LBP variant, called local line binary pattern (LLBP), and Rossi et al. [25] applied it to finger vein recognition, and its accuracy was better than that of LBP and LDP, and it was applied to near-infrared face recognition. The traditional local binary feature extraction method extracts features from each pixel in the image, so the number of extracted features is large, and the information contained in the features is redundant, and the dimension reduction operation is not carried out in the feature extraction process. In order to solve the problem of traditional LBP dimension, center-symmetric local binary pattern (CS-LBP) is proposed in [26], which uses centrosymmetric idea to encode images. The characteristic dimension of CS-LBP is only 1 / 8 of LBP and the processing speed is faster than that of LBP. However, CS-LBP can not describe the large-scale macro structure, and few literatures directly use CS-LBP in finger vein feature extraction. In general, the method based on local binary features does not need to segment the image. Local features are extracted from all pixels of the image, and then the features at the same position are matched or the local feature histogram of the region is matched. However, most of the local features designed by hand are used in this method, and the feature discrimination is weak, which can not reflect the essential characteristics of the data, and the information is redundant.

To sum up, vein veins need to be segmented based on vein pattern method and feature point matching method, but it is difficult to accurately segment low-quality vein images. The method based on principle of statistical characteristics analysis can not consider the local details of the image, and can not achieve good results in finger vein recognition. At present, finger vein recognition methods based on local binary features mostly use hand-designed local features, which have weak discrimination and can not reflect the essential characteristics of data. And the dimension is too high, the algorithm is complex and the processing speed is slow.

Inspired by the appeal work and the idea of centrosymmetric coding, in order to avoid the impact of data redundancy and segmentation on the recognition accuracy, this paper proposes a recognition method based on generating code as shown in Figure 1. In this work, we use block based method to obtain biometric data and generate feature matrix. The feature matrix uses centrosymmetric coding to generate codewords. The data set is divided into training set and test set. Selecting the appropriate number of templates, the minimum Hamming distance between the image and all the templates is taken as the matching score for recognition. Experiments on two public finger vein databases verify the effectiveness of our method. In general, the proposed method improves the performance of finger vein recognition, and promotes the application and development of finger vein recognition to a certain extent.

In this work, we have addressed few research challenges in a succession of our proposed approach. We present an approach to extract feature codewords from finger vein images with low quality, noise and rotation variation. Compared with the traditional methods, our method does not need complex segmentation algorithm, does not face the problem of high feature dimension, avoids the redundancy of information, and has robustness to local changes. Our approach is simple and fast without compromise the accuracy of the result compared to the traditional approach.

The rest of the paper is organized as follows. Proposed approach is discussed in Section 2. In Section 3, we present experimental results. The analysis of our approach is presented in Section 4. Finally, Section 5 concludes the paper.

## 2. Proposed Approach

In this section, we discuss our proposed approach. An overview of our proposed approach is shown in Fig. 1. The different tasks involved in our approach are discussed in details in the following subsections.

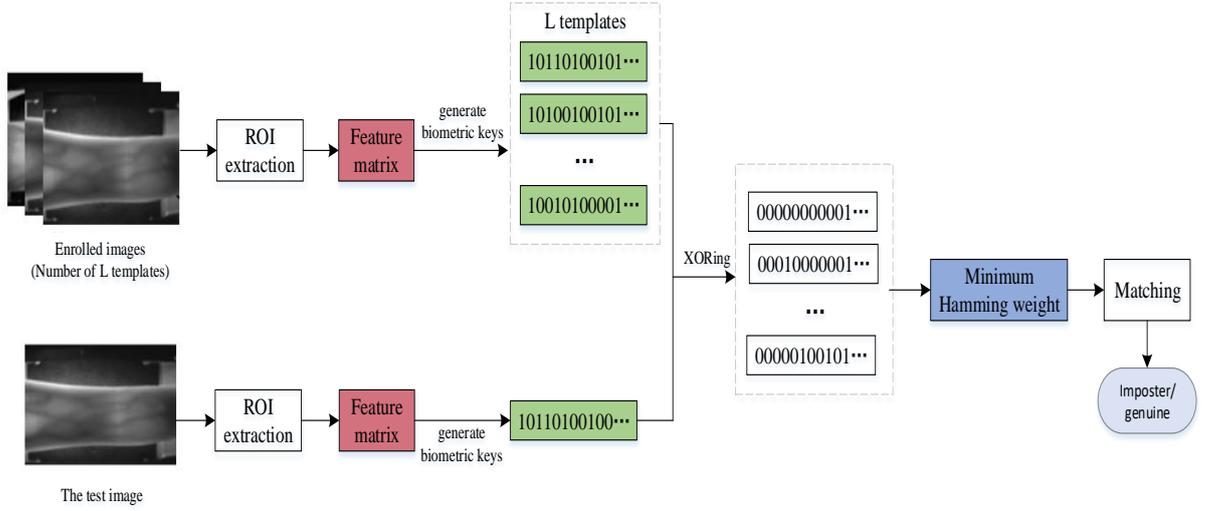

Fig.1. Overall framework of the proposed method

## 2.1 Calculating the block matrix

Our proposed approach is to calculate feature matrix. To do this, we first divide the finger vein image into a small number of blocks. Let $I$ be the finger vein image, then we divide the finger vein image I into a number of small blocks each of size m ×n pixels, m, n<<M, N, where M × N being the size of the input finger vein data. In other words, I can be denoted as a p × q matrix of all blocks as follows (Eq.1).

$$I = \begin{pmatrix} I_{11} & I_{12} & \dots & I_{1q} \\ I_{21} & I_{11} & \dots & I_{11} \\ \dots & \dots & \dots & \dots \\ I_{p1} & I_{p2} & \dots & I_{pq} \end{pmatrix} \qquad (1)$$

Here, $I_{ij}$ denotes the block in $i-th$ row and $j-th$ column in $I$. Note that the blocks, which are on the boundary may not be of equal size, add element 0 to make it equal. For example, the size of the partition block $(m-t) \times (n-l)$ is shown in Fig. 2. Now, we calculate the mean value of each block in this block-based image, and the resulting matrix is the feature matrix.图

$$I_{ij} = \begin{pmatrix} p_{11} & p_{12} & \dots & p_{1(n-l)} \\ p_{21} & p_{22} & \dots & p_{2(n-l)} \\ \dots & \dots & \dots & \dots \\ p_{(m-t)1} & p_{(m-t)2} & \dots & p_{(m-t)(n-l)} \end{pmatrix} \xrightarrow{\text{Add 0 complement}} I_{ij} = \begin{pmatrix} p_{11} & p_{12} & \dots & p_{1(n-l)} & 0 & \dots & 0 \\ p_{21} & p_{22} & \dots & p_{2(n-l)} & 0 & \dots & 0 \\ \dots & \dots & \dots & \dots & \dots & \dots & \dots \\ p_{(m-t)1} & p_{(m-t)2} & \dots & p_{(m-t)(n-l)} & 0 & \dots & 0 \\ 0 & 0 & \dots & 0 & 0 & \dots & 0 \\ \dots & \dots & \dots & \dots & \dots & \dots & \dots \\ 0 & 0 & \dots & 0 & 0 & \dots & 0 \end{pmatrix}_{m \times n}$$

Fig.2. The process that the block size is not enough to make up with 0

## 2.2 generating code

The finger vein code is generated by using the feature matrix. The feature matrix is divided into 3 * 3 small matrices (0 is used to supplement when the boundary is insufficient). The idea of central symmetric encryption is adopted. The central pixel of the small matrix is taken as the central symmetric point, and the pixel value in the field is compared. If it is greater than or equal to 1, otherwise, it is 0. The binary sequence Xi is generated, and the binary sequence of all matrices is connected to form the vein codeword X. The process can be expressed by the following Eq. (2-4):

$$J_i(j) = \begin{cases} 1, n_j \geq n_{j+4} \\ 0, n_j < n_{j+4} \end{cases}, j = 1, 2, 3, 4. \tag{2}$$

$$x_i = \{J_i(1) \| J_i(2) \| J_i(3) \| J_i(4)\}, \tag{3}$$

$$X = \{x_1 \| x_2 \| ... \| x_N\} \tag{4}$$

The pixel value of any point in the image is marked as $n_c$, and the point in its field is marked as $n_j(j = 1, 2, ..., 8)$, $x_i$ is the binary sequence of each small matrix, and X is the binary code generated by the whole graph.

## 2.3 From code to matching:

The minimum Hamming distance is proposed to measure the similarity of the enrolled code and the probe code. Select binary codes of L images for registration as template. The Hamming distance between the probe code and all enrolled codes is calculated, and the ratio of the minimum Hamming distance value to the codeword length is taken as the matching score between the probe code and all enrolled subjects. This method of measuring similarity can reduce the intra-instance distance and increase the inter-instance distance, thereby increasing the recognition rate.

Formally, one probe and one enrolled binary codes are denoted by $X_p^i(i = 1, 2, ..., L)$ and $X_e^j(j = 1, 2, ..., N)$, and $L$, $N$ respectively represent the number of test samples and templates of each finger. Hamming distance can be calculated (Eq. (5,6)):

$$matched(ij) = sum(b_{xor}(ij)) \tag{5}$$

$$b_{xor}(ij) = X_p^i \oplus X_e^j \tag{6}$$

in which $\oplus$ means the boolean exclusive-OR operator. The matching score is defined as Algorithm 1. (Tab.1)

Tab.1 Matching score calculation process

| Algorithm 1 The Matching Score Algorithm |
| --- |
| **Input**: the enrolled binary codes: $\{X_e^j \mid j = 1, 2, ..., N\}$; one probe: $X_p^i$; |
| **Output**: $S_{matching}$ |
| 1: $S_{matching} = 0$ |
| 2: for $j = 1 \rightarrow N$ do |

3:      $S_j = matched(ij)$

4:    end for

5:    $S_{matching} = \dfrac{\min\{S_j \mid j = 1, 2, ..., N\}}{length(X_p^i)}$

# 3. Experiments and Experimental Results

## 3.1 Databases

Two open finger vein databases are used to evaluate the performance of our vein code indexing. As some databases only contain 6 images per finger, we ignore the fingers with 6 images and only use the fingers with 12 images to ensure the consistency of all databases. The details of these databases are given below.

1) HKPU Database [8]: This database is from Hong Kong Polytechnic University, including 3,132 images from 312 fingers. Each of first 210 fingers has 12 images, captured in two separate sessions, and others each has 6 images, captured in one session. All of the images are 8-bit gray level BMP files with a resolution of 513×256 pixels.

2) USM Database[27]: This database is from Universiti Sains Malaysia. It consists of 492 fingers, each with 12 images. These images are 8-bit gray level JPG files with a resolution of 640×480 pixels. The details of two databases we used are listed in Tab. 2.

Tab. 2. The details of databases

| Database | Finger number | Image number per finger | Size of raw image(pixels) | ROI image |
|---|---|---|---|---|
| HKPU | 312 | 6/12 | 513x256 | Extract using the method of document [8] |
| USM | 492 | 12 | 640x480 | Database includes ROI images |

## 3.2 Experimental Protocols

Three experiments are designed here. First, we assess the effect of parameters (i.e., the template number and decision threshold (DT)) on the recognition performance and choose the best value for the following experiments. Second, we compare our method with some typical finger vein recognition methods. Third, we test the potential of our vein code in the matching task by comparing them with some state-of-the-art finger vein matching methods.

The database settings of the recognition task in the experiment are given.

For the open database HKPU, we only consider the fingers with 12 pictures. There are 210 finger samples in total. The USM database has 12 images of all fingers, and the images in the whole database will be used for the experiment. It is observed that values of the size of the block matrix have a very small impact on the performance of the algorithms, and there are differences in image size between the two databases. Therefore, the effect of block size on the recognition rate is not is not listed in the result. In the experiment, select the appropriate block size directly; for HKPU and USM data, the block size is 3*8 and 5*5, respectively. The matching performance is evaluated by the equal error rate (EER) and the recognition rate [33]. The EER is the error rate when the false acceptance rate (FAR) is equal to the false rejection rate (FRR). We use the intra-instance (1:1) and inter-instance(1:N) as the main indicators to measure the test identification.

## 3.3. Selection of Parameter Value

In this experiment, we study the performance of our method with different parameter values. The parameters include the number of template and decision threshold (DT). This experiment is conducted on the most popular HKPU database and USM database. We fix the DT when considering the influence of different template numbers on the recognition rate, and vary the template number from 2 to 8 with 2 as the interval. When the number of templates is N, the first N images of the database are selected as the template, and the remaining images are used as the test samples. The recognition performance are listed in Tab.3.

Tab.3. The recognition performance under different number of templates

| database | The template number(N) | 2 | 4 | 6 | 8 |
|----------|------------------------|------|------|------|------|
| HKPU | EER | 6.85 | 5.52 | 2.89 | 2.91 |
| USM | EER | 2.35 | 2.08 | 1.16 | 0.92 |

From the results, we can see that the recognition performance improves with the increasing template number. It can be easily understood that larger number of templates can more powerfully extract details of finger vein patterns, which is helpful for the recognition performance. In addition, the results also tell us that the improvement of the recognition performance is not significant when the template number varies from 6 to 8. Considering the convenience of image collection in practical application, we use 6 template in following experiments.

We also fix the template number at 6 and vary the number of matching score from 0.18 to 0.21 with 0.11 as the interval to get the decision threshold (DT). The recognition rate of DT under different values is presented in Tab.4.

Tab.4. The recognition rate under different DT values

| HKPU database | The decision threshold | intra-instance (1:1) | | | inter-instance(1:N) | | |
|---------------|-----------------------|------------|-------------|-------------------|-------------|-------------|-------------------|
| | DT values | Total times | False times | Recognition rate | Total times | False times | Recognition rate |
| | 0.18 | 1260 | 55 | 95.6 | 263340 | 2465 | **99.1** |
| | 0.19 | 1260 | 43 | 96.6 | 263340 | 5228 | 98.0 |
| | 0.20 | 1260 | 34 | 97.3 | 263340 | 9899 | 96.2 |
| | 0.21 | 1260 | 29 | 97.7 | 263340 | 16911 | 93.6 |
| | 0.18 | 2952 | 83 | 97.1 | 1449432 | 2000 | **99.7** |
| USM database | 0.19 | 2952 | 57 | 98.1 | 1449432 | 5127 | **99.6** |
| | 0.20 | 2952 | 39 | 98.7 | 1449432 | 11263 | **99.2** |
| | 0.21 | 2952 | 32 | **98.9** | 1449432 | 20730 | 98.7 |

From the results, we can see that there is a critical decision threshold (DT) value. If this threshold value is exceeded, the 1:1 recognition rate rises and the 1:n recognition rate decreases. This threshold is different for different finger vein libraries. This means that different finger vein recognition devices need to adjust the threshold to achieve the best recognition status. For clarity, we will use the equal error rate (ERR) indicator to compare the performance of different algorithms.

### 3.4 Comparision with existing finger vein recognition methods

In this part, we investigated the accuracy of our method in verification mode compared with the various existing finger vein recognition methods, i.e., LBP [20], LLBP [24], ELBP[11], PLBP

[19], Repeated line tracking (RLT)[5], Mean curvature(MC) [7], Maximum curvature point (MCP) [6], the curvature in radon space (CRS)[9], Gabor [8], CNN[1]and anatomy structure analysis based vein extraction (ASAVE) [3].

The comparison is performed on HKPU databases. The performance is measured by the equal error rate (EER). Tab. 5 shows the EER of different methods on HKPU databases. In the following, we will analysis these results.

First, our proposed method compare with the algorithm without segmentation(e.g., LBP [16], LLBP [13], PLBP,ELBP). Compared with these methods, our method can extract the macro and micro features of the image at the same time, express the image information completely, and enhance the classification effect. And the algorithm is more simple, insensitive to local small changes, and has stronger robustness to noise.

Second, comparison with the algorithm with segmentation (e.g., RLT[5], MCP[6], ASAVE [3] , etc.), our proposed method achieves better performance on the HKPU databases. The method that needs segmentation algorithm is greatly affected by the image quality and requires high image quality. For instance, ASAVE is a vein pattern based method that utilizes the anatomy structure and imaging characteristic of vein patterns. The vein pattern based method does not work well due to the presence of several low quality images in open databases, such as the limited vein pattern in some images and the low contrast between the finger vein and non-vein pattern region. While these low quality images belonging to same class have the similar gray distribution, our proposed method can explore the gray distribution of images, and achieves better performance.

Tab.5.Compare the recognition performance of our proposed method and existing methods in HKPU database

| method | algorithm | EER |
|---|---|---|
| No segmentation algorithm required | LBP | 18.8 |
| | LLBP | 15.77 |
| | PLBP | 7.96 |
| | ELBP | 5.59 |
| Need segmentation algorithm | Repeated line tracking | 16.31* |
| | Mean curvature | 4.03 |
| | Maximum curvature point | 18.99* |
| | the curvature in radon space | 2.96 |
| | Gabor | 4.61* |
| | CNN | 3.02* |
| | ASAVE | 2.91* |
| | our proposed method | 2.89 |

*Cited from [2]

## 3.5 Comparison with state-of-the-art  finger vein indexing method

Recently, Yang et al. [2] proposed a vein pattern integration framework in which finger vein indexing and finger vein matching are both considered, and called it as weighted vein code indexing. And Hu et al. [18] used block multi-scale uniform local binary pattern features and block two-directional two dimension principal component analysis to verify vein.

In this section, we compare our proposed method with them, and EER is used as benchmark. The compared results are shown in Tab.6. From Tab.6, we can find our proposed method performs better than weighted vein code indexing [2]. This is mainly because there are several low quality images in HKPU databases. Low-quality images have limited vein patterns and low contrast between the finger veins and the non-vein pattern region. These characteristic of image is harmful for methods using vein pattern (Such as weighted vein code indexing [2]). While our proposed method can overcome the problem of low-quality images. Therefore, our method outperforms the weighted vein code indexing at the EER. Although our proposed method belongs to finger vein recognition methods, it is most efficient than the indexing methods (such as weighted vein code indexing).

Tab.6 Comparison the recognition performance with state-of-the-art finger vein method

| Databases | | EER | Years of publication |
|---|---|---|---|
| HKPU database | Weighted vein indexing [2] | 3.33 | 2019 |
| | our proposed method | 2.89 | —— |
| USM database | block multi-scale uniform local binary pattern features [18] | 1.89 | 2020 |
| | our proposed method | 1.16 | —— |

In addition, We also compare the recognition between our proposed method and Literature [18] in USM databases. We can see that our method has better recognition performance than literature [18]. That's because the method in literature[18]only considers the local information of the image, while our method takes into account both the local macro and micro information of the image, and is robust to small local changes, so it has stronger anti-noise performance than reference [18]

In summary, in above comparison about the EER, our proposed method achieves promising performance on open databases over the method whether it needs to be segmented or not.

# 4 Conclusions and future work

Existing finger vein finger vein recognition methods are not satisfactory with the recognition performance. Algorithms that need to segment images (such as maximum curvature, repeated linear tracking, Gabor filtering, ASAVE, etc.) have high requirements for image quality and are not practical. The method without segmentation algorithm (such as LBP) is computationally complex and has large data redundancy, so it is not effective to directly use it in finger vein recognition. The proposed finger vein code generation algorithm is simple in calculation, does not need complex segmentation algorithm, can overcome the problem of low image quality, and has stronger robustness to image noise.

We conduct extensive experiments on two finger vein databases for testing our method. Experimental results indicate that our method can outperform most state-of-the-art methods and has competitive potential to perform the matching task. There is one worthy direction for future work, i.e., collecting a large-scale finger vein database. Currently, there is no publicly available large-scale database for the performance comparison of finger vein recognition methods. The image size, the captured finger region, the gray level, and the background noise are varied from one database to another; this is not helpful for fairly evaluating the indexing performance.